# Towards Better Surgical Instrument Segmentation in Endoscopic Vision: Multi-Angle Feature Aggregation and Contour Supervision

Fangbo Qin, Shan Lin, Yangming Li, Randall A. Bly, Kris S. Moe, Blake Hannaford*, *Fellow, IEEE*

*Abstract*—Accurate and real-time surgical instrument segmentation is important in the endoscopic vision of robot-assisted surgery, and significant challenges are posed by frequent instrument-tissue contacts and continuous change of observation perspective. For these challenging tasks more and more deep neural networks (DNN) models are designed in recent years. We are motivated to propose a general embeddable approach to improve these current DNN segmentation models without increasing the model parameter number. Firstly, observing the limited rotation-invariance performance of DNN, we proposed the Multi-Angle Feature Aggregation (MAFA) method, leveraging active image rotation to gain richer visual cues and make the prediction more robust to instrument orientation changes. Secondly, in the end-to-end training stage, the auxiliary contour supervision is utilized to guide the model to learn the boundary awareness, so that the contour shape of segmentation mask is more precise. The proposed method is validated with ablation experiments on the novel Sinus-Surgery datasets collected from surgeons' operations, and is compared to the existing methods on a public dataset collected with a da Vinci Xi Robot.

## I. INTRODUCTION

WITH the rapid development of surgical robotics, computer-assisted intervention, and computer vision technologies, increasing attention has been paid to how intelligent endoscopic vision can benefit surgery performance and clinical outcome [1-3]. For example, endoscopic video-based navigation was used for registration in sinus surgery [4], the dense reconstruction of 3D surgical scene was developed for handheld monocular endoscopy [5], the endoscopic video-based augmented reality (AR) could increase surgeon's visual awareness of high-risk targets [6], and the operative skill assessment was realized based on surgical video [7]. Surgical robots, such as the widely-applied da Vinci systems [8] and the collaborative research platform Raven-II [9], can gain more flexibility and autonomy from endoscopic vision. In [10], real-time endoscopic vision was used to guide semi-autonomous tumor resection. The real-time 3D tracking of articulated tools helped the safe tool-tissue interaction of da Vinci robot [11]. For the above endoscopic vision perception applications, instrument segmentation, i.e. separating the instrument from the tissue background, is a key technology, providing the essential region information so that we can estimate the instrument position and analyze the instrument and tissue separately.

Surgical scene complexity is usually more limited than natural scenes, which involves only several types of instruments and organs. However, frequent instrument-tissue contacts, non-uniform illumination, deformable reflective surfaces, and continuous change of observation perspective pose many challenges to image segmentation.

### A. Related Works

In recent years, deep neural networks (DNNs) have significantly advanced endoscopic image perception [12]. With enough training data, DNNs surpass the traditional handcrafted feature based methods due to DNN's feature learning ability and deep hierarchical structure. For real-time instrument segmentation, the lightweight ToolNet models were designed based on holistically-nested structures, and outperformed the classical segmentation model FCN-8s [13, 14]. Laina *et al*. combined the landmark localization and segmentation tasks in one DNN model [15]. The recurrent neural networks layers were embedded within the convolutional model to model the pixel interdependencies [16]. In [17], the ToolNet-C model was a cascade of a feature extractor and a segmentor, which were trained on numerous unlabeled images and a few labeled images, respectively. U-net is a classical architecture proposed for biomedical images, which has a symmetric structure and multiple skip connections [18].

Facing more challenging surgical scenes, the DNN-based segmentation requires larger model capability and deeper structure. Shvets *et al*. modified the LinkNet [19] and TernausNet [20] models to LinkNet-34 and TernausNet-16 for instrument segmentation, respectively, between which

Manuscript received: February 24, 2020; Revised May 21, 2020; Accepted June 19, 2020.This paper was recommended for publication by Editor Eric Marchand upon evaluation of the Associate Editor and Reviewers' comments.
This work was supported by National Science Foundation (IIS-1637444), and National Natural Science Foundation of China (61703398).
F. Qin is with Research Center of Precision Sensing and Control, Institute of Automation, Chinese Academy of Sciences, Beijing 100190, China. {qinfangbo2013@ia.ac.cn}
S. Lin and B. Hannaford are with Department of Electrical Engineering, University of Washington (UW), Seattle, WA 98195-2500, USA. R. A. Bly, and K. S. Moe are with Department of Otolaryngology-Head & Neck Surgery, UW, Seattle, USA. {corresponding author: blake@uw.edu}
Y. Li is with Department of Electrical Computer and Telecommunications Engineering Technology, Rochester Institute of Technology, USA.
Digital Object Identifier (DOI): see top of this page.

TernausNet-16 provided the better accuracy [21]. Islam *et al.* proposed a real-time instrument segmentation model with a multi-resolution feature fusion module, which fuses the two high- and low-resolution feature maps given by the main and auxiliary branches, respectively [22]. In the experiments, the proposed multi-resolution method outperformed the general-purpose semantic segmentation models like PSPNet [23] and ICNet [24]. Ni *et al.* designed the LWANet based on a lightweight encoder and channel-attention guided decoder, which achieves real-time speed for large surgical images [25]. It is advantageous to incorporate optical flow based inter-frame motion cues into segmentation [26]. However, the optical flow estimation task could be challenging in endoscopic vision, especially when the illumination and background are highly dynamic.

*B. Motivation and Contributions*

This paper aims to propose embeddable methods to realize relative accuracy improvement for the current surgical instrument segmentation models. We observe the two problems: 1) Although the random rotation with large-range is usually included in the data augmentation during training, the segmentation could still be sensitive to the instrument orientation's change. In endoscopic surgery scenarios, the endoscope's axial symmetry and the dexterous motion frequently lead to instrument rotation in the endoscopic view. 2) The predicted segmentation masks are prone to have incomplete and inaccurate parts near the instrument boundary. Considering these two problems, our contribution is as follows.

1) The *multi-angle feature aggregation* (MAFA) method is proposed, which can be flexibly incorporated with a DNN segmentation model, without increasing the model's parameter number. The aggregation of visual features augmented by multi-angle rotation provides better robustness and accuracy.

2) The *contour supervision* is utilized as an auxiliary learning task in the training stage, which guide the model to learn features highly related to instrument boundary, so that the prediction near the boundary can be refined.

3) The novel *Sinus-Surgery datasets* (online available[1]) were collected from endoscopic sinus surgeries performed by surgeons, which are featured by dexterous tip motion, narrow operation space and close lens-object distance.

II. METHODS

*A. Semantic Consistency under Rotation-and-Alignment*

For endoscopic images, we assume the semantic consistency under rotation-and-alignment. The *rotation* operation is to rotate a map with an angle around the image center, and the *alignment* operation is to rotate the map with the same angle but in the reversed direction. Intuitively, if we actively rotate an image with a certain angle, view the rotated image, and then align the image back to the original angle, the semantics we get should be consistent with those we directly get from the un-rotated image. As shown in Fig 1, a pixel on an endoscopic image $\mathcal{I}$ is labeled by the position $p$, which belongs to the instrument region.

[1]https://github.com/SURA23/Sinus-Surgery-Endoscopic-Image-Datasets

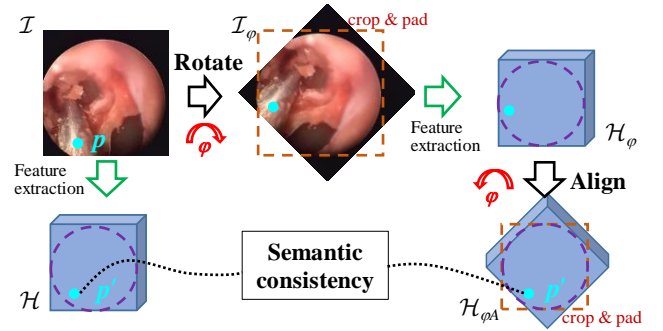

Fig. 1. Semantic consistency under rotation-and-aligment. The purple dashed circle indicates the visible circle in endoscopic image, beyond which is the black border. The cyan dot marks the same point on an instrument.

Firstly, after a *rotation* operation with the angle $\varphi$, the rotated image is given by,

$$\mathcal{I}_\varphi = Rot(\mathcal{I}; \varphi) \quad (1)$$

The pixel position $p$ is shifted by the rotation operation.

The feature extraction implemented by convolutional layers extracts the feature map $\mathcal{H}$ from the input $\mathcal{I}$, which is expressed as $\mathcal{H}=F(\mathcal{I})$. The rotated image $\mathcal{I}_\varphi$ is input to the shared feature extraction module and converted to $\mathcal{H}_\varphi=F(\mathcal{I}_\varphi)$.

Secondly, the *alignment* operation is used to rotate $\mathcal{H}_\varphi$ in the reversed direction by the same angle $\varphi$, namely,

$$\mathcal{H}_{\varphi A} = Rot(\mathcal{H}_\varphi; -\varphi) \quad (2)$$

Thus the feature map $\mathcal{H}_{\varphi A}$ is aligned to $\mathcal{H}$ in orientation. The image rotation relies on the cropping of out-of-field pixels and zero-padding near image corners, as is shown in Fig.1. The pixels within the visible circle of endoscope are not affected by the cropping and padding.

Thirdly, let $p'$ indicate the feature point on $\mathcal{H}$ corresponding to the pixel $p$ on the raw image $\mathcal{I}$, the assumption of semantic consistency under rotation-and-alignment requires that the feature $\mathcal{H}_{\varphi A}(p')$ has the consistent semantics with the feature $\mathcal{H}(p')$ at the same position $p'$, although generally these two feature vectors are not equal in values. Here the semantic consistency means that the features in the same channel refer to the same attribute of object, so that they can be aggregated together in a channel-wise manner for object perception.

*B. Multi-angle Feature Aggregation*

We propose to generate augmented features by the rotation-and-alignment operations given one input image $\mathcal{I}$ and multiple rotation angles $\{\varphi_k\}$, as expressed by,

$$\mathcal{H}_{\varphi A}^{(k)} = Rot\left(F\left(Rot(\mathcal{I}; \varphi_k)\right); -\varphi_k\right) \quad (3)$$

where $k=1,2,\ldots,N_A$ and $N_A$ is the number of rotation angles. In the set $\{\mathcal{H}_{\varphi A}^{(k)}\}$, a different $\mathcal{H}_{\varphi A}^{(k)}$ contains the consistent semantic feature when observing the input image from a different angle $\varphi_k$. The choice of $N_A$ is a tradeoff between feature augmentation and computation load. A larger $N_A$ leads to richer visual features but longer runtime.

The aggregated feature is produced by averaging the multi-angle feature maps,

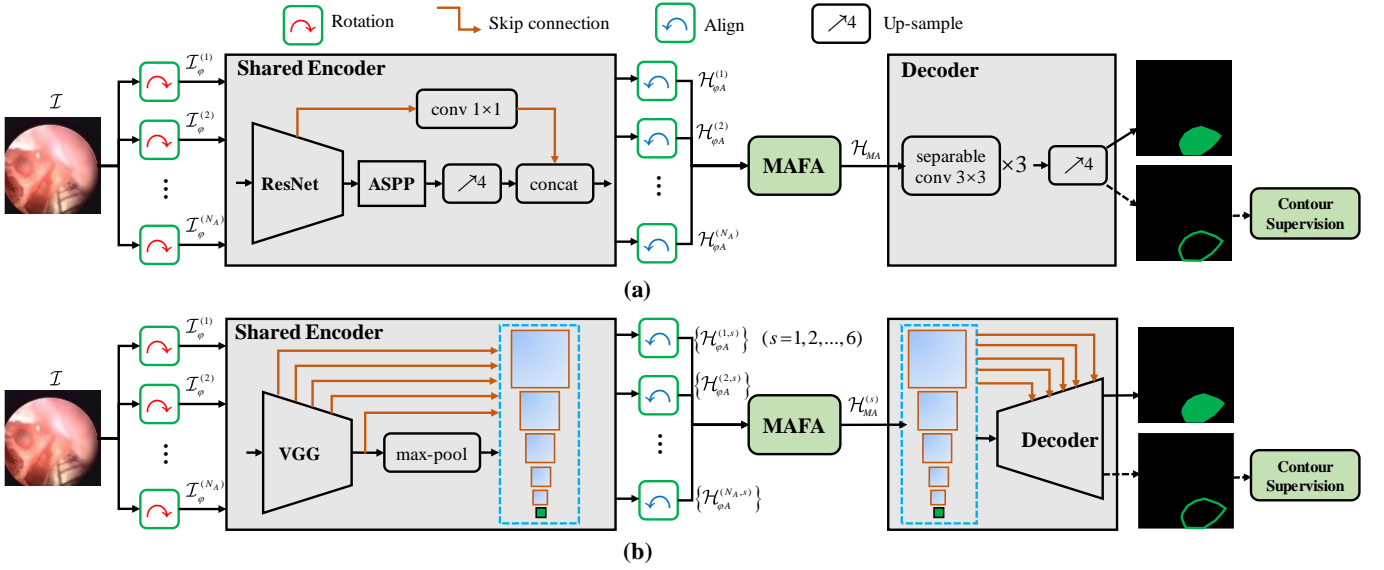

Fig. 2. Combination of MAFA and contour supervision with deep segmentation models. (a) Combine with DeepLabV3+ architecture. (b) Combine with TernausNet-16 architecture. The dashed lines before the output maps mean that the contour prediction is optionally used in the inference stage.

$$\mathcal{H}_{MA}(i,j,c) = \frac{1}{N_A} \sum_{k=1}^{N_A} \mathcal{H}_{\varphi A}^{(k)}(i,j,c) \quad (4)$$

where $i$, $j$ and $c$ are the indexes in height, width and channel of the feature map, respectively. The contribution of each rotation is considered the same in the averaging operation. Thus, $\mathcal{H}_{MA}$ has the same feature channel number with $\mathcal{H}_{\varphi A}^{(k)}$ but contains the aggregated information generated by multi-angle rotation-and-alignment.

### C. Contour Supervision and Loss Function

Inspired by [27], the multi-task learning supervised by both region segmentation and boundary prediction can make the model pay more attention to the object contour. The contour is informative because it not only localizes the precise edges but also represents the object's outer shape. Therefore, the final outputs of the segmentation model are two maps given by softmax function: the segmentation map $\mathcal{S}$ and the contour map $\mathcal{C}$, whose sizes are the same as the input image. In our binary segmentation task, both the maps have 2 channels. The ground truth of contour map is $\mathcal{C}'$, obtained by finding the outer contour of the foreground regions in the ground truth of segmentation map $\mathcal{S}'$. The contour width is empirically set as 3 pixels in this work.

As is different with [27], to increase contour sharpness, we use the Dice loss instead of cross-entropy loss to supervise the learning of contour prediction, as expressed by,

$$\mathcal{L}_C = 1 - \sum_{i,j,k=2} 2\mathcal{C}_{i,j,k}\mathcal{C}'_{i,j,k} \Big/ \left( \sum_{i,j,k=2} \mathcal{C}^2_{i,j,k} + \sum_{i,j,k=2} \mathcal{C}'^2_{i,j,k} + \tau \right) \quad (5)$$

where $i$, $j$ and $k$ are the indexes in height, width and channel. Because only contour pixels are used to calculate Dice loss, the summation in (5) only involves $k=2$ that indicates the contour probability channel. $\tau$ is a small positive constant like $1\times e^{-6}$ to stabilize the calculation. The segmentation learning is based on the standard cross-entropy loss,

$$\mathcal{L}_S = -\frac{1}{N} \sum_{i,j,k} \mathcal{S}'_{i,j,k} \log(\mathcal{S}_{i,j,k}) \quad (6)$$

where $N$ is the count of all the map elements. Finally, the multi-task learning is supervised the combination of these two losses,

$$\mathcal{L} = \mathcal{L}_S + \mathcal{L}_C \quad (7)$$

## III. EXAMPLES

The proposed MAFA and contour supervision methods can be flexibly combined with a segmentation model. In this section, two segmentation models, DeepLabv3+ [28] and TernausNet-16 [21], are used as examples to describe the integration of proposed methods. These two models are both based on encoder-decoder architecture, as shown in Fig. 2. The former is a typical model with spatial pyramid pooling intermedia and sparse skip connection. The latter is a typical model with the U-shape and dense skip connections.

After integrating MAFA, an input image $\mathcal{I}$ is rotated by $N_A$ angles, $\{\varphi_k\} = \{2\pi \times (k-1)/N_A\}$ ($k=1,2,\ldots,N_A$), which cover the rotation range [0°, 360°] with the equal interval. This is a general configuration assuming that the instrument orientation can be arbitrary. All the rotated images $\mathcal{I}_{\varphi}^{(k)}$ are processed by a shared encoder in parallel, thus the GPU's parallel computation advantage is leveraged. The encoder outputs are aligned to the original angle, so that the multi-angle feature maps $\mathcal{H}_{\varphi A}^{(k)}$ are produced. Then, the MAFA block fuses the multiple maps as one aggregated feature map, as is described by Eq. (4). Finally, the decoder converts the aggregated feature map to the segmentation map and contour map.

### A. DeepLabv3+ with MAFA and Contour Supervision

DeepLabv3+ model is mainly characterized by its atrous spatial pyramid pooling (ASPP) block and brief structure with only one skip connection route. We customize the original DeepLabv3+ model as described below.

1) The backbone feature extractor converts the input image to a feature map with the *output-stride* of 16. Output-stride is the ratio of input image resolution to map resolution. If using ResNet-50 [29] as backbone, the output has 2048 channels.

2) The ASPP block has four branches: an image pooling branch, a 1×1 convolution branch and two 3×3 convolution branches with the atrous rates {2, 4}. Each branch's output map has $N_{high}$=256 channels. The four branches' outputs are concatenated along channels and then compressed as a $N_{high}$-channel map by a 1×1 convolution layer.

3) The skip connection pulls out the low-level feature map, e.g. the output of the 8th layer in ResNet-50. The low-level feature map is adapted to $N_{low}$=32 channels by 1×1 convolution. The 256-channel high-level feature map is resized to 4× larger with bilinear-interpolation, and concatenated with the low-level feature map. Thus the final encoder output is a multi-scale feature map with ($N_{high}$+$N_{low}$) channels.

4) In the decoder, the three depthwise separable convolution layers have 3×3 kernel and 1×1 stride. The first two layers have the 128-channel outputs, and the last layer has the 4-channel output with the 4 output-stride. The first two channels are for segmentation and the last two for contour prediction. Finally, the 4× up-sampling with bilinear interpolation is used to resize the maps up to the input size. Batch normalization is applied after convolutional layers to stabilize the training, and ReLUs are used for activation.

### B. TernausNet-16 with MAFA and Contour Supervision

TernausNet-16 is featured by its pre-trained VGG-16 encoder and its decoder block design. The original TernausNet-16 model is combined with MAFA without customization. Because TernausNet-16 has as many as 5 skip connections, we define the encoder output as a set of 6 feature maps: one bottleneck feature map and five skipping feature maps, as shown in Fig. 2. Note that the 6 feature maps have different sizes, encoding hierarchical visual cues.

With the multi-angle rotation, the encoder outputs include 6×$N_A$ feature maps, which are aligned to the original angle and labeled as $\{\mathcal{H}_{\varphi_A}^{(k,s)}\}$ ($k$=1,2,…,$N_A$; $s$=1,2,…,6). $k$ and $s$ indicate the angle index and scale index of the feature maps, respectively. MAFA block fuses the multi-angle hierarchical maps $\{\mathcal{H}_{\varphi_A}^{(k,s)}\}$ to the aggregated hierarchical maps $\{\mathcal{H}_{MA}^{(s)}\}$, which are input to the decoder in sequence. ReLUs are used for activation and softmax is used as the output function.

## IV. EXPERIMENTS AND RESULTS

We introduce the two novel endoscopic image datasets collected from sinus surgeries performed by surgeons, with which a series of experiments were carried out to investigate the effectiveness of the proposed methods. In addition, we compared the performances of different models based on a public dataset collected with a da Vinci Xi robot.

### A. Sinus-Surgery Datasets

1) *Sinus-Surgery-C* dataset: It was collected from 10 surgeries conducted on 5 cadaver specimens. Each cadaver specimen was operated on twice, through the left and right nasal cavities, respectively. 9 surgeons participated in the experiments using the same micro-debrider instrument. Each surgeon operated on one specimen, except one surgeon who operated on two. Cadaver specimens had apparent tissue color differences. Thus, the 10 videos contained diversity in both tissue backgrounds and surgical motion skills. As is shown in Fig. 3(a), The endoscopic videos were recorded with the 320×240 resolution and 30fps frame rate, utilizing the Stryker 1088 HD camera system and the Karl Storz Hopkins ⌀4mm 0° endoscope. The video durations ranged from 5 minutes to 23 minutes. From each video an image set was extracted with the equal time interval of 2 seconds. All the images were center-cropped to 240×240 and manually given ground truths of instrument foreground. Finally, we obtained 10 subsets whose image numbers were 285, 532, 344, 162, 491, 688, 406, 291, 536, and 610, respectively. The total image number was 4345, among which 3606 images contained instrument.

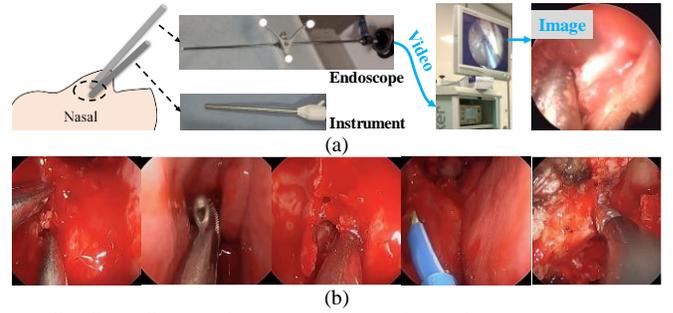

Fig. 3. Sinus-Surgery datasets collection. (a) Endoscopic sinus surgery schematic and a sample image of cadaver surgery. (b) Sample images of multiple instruments used in live surgeries.

2) *Sinus-Surgery-L* dataset: It was collected from 3 live operating room surgeries conducted on 3 patients. The total duration of the 3 endoscopic videos was about 2.5 hours. The dataset was built in a similar way with Sinus-Surgery-C. The 3 subsets had 1154, 2801, and 703 images, respectively. The total image number was 4658, among which 4344 images contained instrument. Compared to Sinus-Surgery-C, this dataset was more challenging because there were multiple instrument types, sometimes inadequate visualization and bodily secretions altering the scene, as is shown in Fig. 3(b).

### B. Training

The deep models were trained with the Adam optimizer, whose exponential decay rates of the 1st and 2nd order moment estimates were 0.9 and 0.999, respectively. The training epoch and batch size were 60 and 16, respectively. The learning rate was initialized as 0.0005. The exponential decay strategy of learning rate was used, whose decay rate and step were 0.5 and 15 epochs, respectively. The backbone networks were pre-trained on ImageNet, and their outputs passed through a dropout layer with 0.5 keep-rate to avoid overfitting. Data augmentation is beneficial for generalization ability. Before each optimization step, the random augmentation was applied including: hue, brightness, saturation, contrast, left-right flip, up-down flip, rotation, zoom in with random center point, zoom-out and zero-padding. *Note that the random rotation angles ranged from 0° to 360°.* The hardware configuration included a 3.70GHz Intel i7-8700K CPU and two NVIDIA RTX2080ti GPUs.

## C. Evaluation

*1) Metrics*: Besides the standard metrics of Dice similarity coefficient (DSC) and intersection of union (IOU), we designed three additional metrics: *rotational mean of IOUs* (RM$_{IOU}$), *rotational standard-deviation of IOUs* (RSD$_{IOU}$), and *IOU near boundary* (IOU$_{NB}$), as given by,

$$\text{RM}_{IOU} = mean\left(\{IOU_i\}_{i=1,2,\ldots,6}\right), \text{RSD}_{IOU} = stdev\left(\{IOU_i\}_{i=1,2,\ldots,6}\right),$$

where IOU$_i$ here is the IOU of the rotated ground truth and the prediction of the rotated input image, given the $i^{th}$ rotation angle. The six angles are {0°, 60°, 120°, 180°, 240°, 300°}. The *mean* and *stdev* function is used to get the mean and standard deviation of the set {IOU$_i$}.

$$\text{IOU}_{NB} = \frac{|(S \cap G) \cap B|}{|(S \cup G) \cap B|},$$

where | | is the counting operation. *S* an *G* are the foreground pixels of the prediction and ground truth, respectively. *B* denotes the near-boundary mask, whose foreground pixels form a 20pixel-width band along the instrument boundary.

Thus, RM$_{IOU}$ and RSD$_{IOU}$ are rotation-invariance metrics. The less deviation between RM$_{IOU}$ and IOU, also the smaller RSD$_{IOU}$, indicate that the segmentation behavior is more consistent under rotations. IOU$_{NB}$ is a metric of contour accuracy, which focused on the pixels near the boundary. Over all the testing examples, the mean values of these metrics are calculated as mDSC, mIOU, mRM$_{IOU}$, mRSD$_{IOU}$ and mIOU$_{NB}$, respectively. Note that the threshold changing probability map to binary map was 0.5.

*2) K-fold Cross-validation*: To investigate the generalization ability and avoid bias caused by dataset split, we used *K*-fold cross-validation with the Sinus-Surgery datasets. Considering the scale of dataset and the number of experiments, we set *K*=3. For Sinus-Surgery-C dataset, the three folds were formed by its 1$^{st}$-4$^{th}$, 5$^{th}$-7$^{th}$, and 8$^{th}$-10$^{th}$ subsets, respectively. For Sinus-Surgery-L dataset, the three folds were formed by its 1$^{st}$, 2$^{nd}$, and 3$^{rd}$ subsets, respectively. For each *K*-fold cross-validation experiment, the mean and standard-deviation of the K scores were calculated and reported in Table I and II.

## D. Comparison Experiments on MAFA

To validate the effectiveness of MAFA and compare the different configurations, a series of experiments were carried out on Sinus-Surgery-C dataset using the DeepLabV3+ model described in Section III.A, whose backbone was the pretrained MobileNet [30] (the width multiplier was 1.0), and the results are reported in Table I.

*1) Influence of rotational data augmentation*: Rows 1 and 2 in Table I show the results of the models trained *without* and *with* the data augmentation via random rotation, respectively. The rotational data augmentation provided a slight accuracy improvement and a significant increase on the rotation-invariance related metric mRM$_{IOU}$ by 10.3%.

*2) Choice of rotation number*: Rows 3 to 7 in Table I show the results of the models with MAFA and different rotation number choices. As $N_A$ was increased from 2 to 6, mRM$_{IOU}$ increased gradually and was close to mIOU, meanwhile mRSD$_{IOU}$ decreased gradually, which demonstrated that a larger $N_A$ could increase the rotation-invariance. The inference runtime became larger as $N_A$ increased.

TABLE I
COMPARISON EXPERIMENT RESULTS ON MAFA
WITH DEEPLABV3+ (MOBILENET) AND SINUS-SURGERY-C DATASET

| MAFA config. | mDSC (%) | mIOU (%) | mRM$_{IOU}$ (%) | mRSD$_{IOU}$ (%) | Time (ms) |
|---|---|---|---|---|---|
| 1. × (no rot.) | 85.2±1.4 | 79.5±1.4 | 60.3±4.7 | 25.7±3.9 | 3.2 |
| 2. × | 85.7±2.4 | 80.0±3.1 | 70.6±1.0 | 22.9±1.4 | 3.2 |
| 3. $N_A$=2 | 88.5±1.0 | 83.4±0.8 | 83.6±0.4 | 4.7±0.1 | 4.6 |
| 4. $N_A$=3 | 89.9±0.7 | 85.0±0.9 | 84.5±0.7 | 4.2±0.9 | 5.4 |
| 5. $N_A$=4 | 89.7±0.6 | 85.0±0.9 | 84.9±0.4 | 3.6±0.6 | 6.2 |
| 6. $N_A$=5 | 89.6±1.2 | 84.9±1.5 | 85.0±1.0 | 3.3±0.4 | 7.4 |
| 7. $N_A$=6 | **90.2**±0.9 | **85.6**±1.2 | **85.5**±0.9 | **1.6**±0.1 | 8.3 |
| 8. max-out | 89.1±0.9 | 84.3±1.1 | 84.3±1.0 | 3.3±0.4 | 6.2 |
| 9. mid-layer | 84.5±3.8 | 78.9±4.6 | 73.2±1.3 | 18.7±1.4 | 4.1 |
| 10. last-layer | 89.5±0.8 | 84.6±1.1 | 83.7±0.6 | 4.8±0.9 | 4.4 |
| 11. ensemble | 85.0±1.1 | 78.7±1.4 | 78.4±1.5 | 3.4±0.0 | 5.3 |

\* The bold fonts indicate the best performance in the column.

*3) Other configurations*: Rows 8 to 11 in Table I show the results of the trials with other configurations. In all four experiments, $N_A$ was set as 4. First, we replaced the feature average aggregation with the feature *max-out* aggregation in MAFA. Max-out means to choose the max value among the $N_A$ inputs as the output. Comparing rows 5 and 8, we found that the average aggregation performed better than the max-out aggregation. Second, we tried to align and aggregate the multi-angle features earlier, namely after *the middle (11$^{th}$) and the last (27$^{th}$) layers* of the backbone. Comparing rows 9-10 with row 5, we found that the runtime was lower while the accuracies decreased, after using MAFA earlier. Specially, if MAFA was put in the middle of backbone, the inner structure of backbone was changed and did not match the pretrained weights, so that the training result was degraded. Finally, we tested the *multi-angle ensemble* technique, namely, we rotated the inputted image also with the same $N_A$ angles, then input these images into the model of row 2 and averaged the $N_A$ output probability maps before thresholding. Its result was worse than that of MAFA.

## E. Ablation Experiments on MAFA and Contour Supervision

In the ablation experiments with the Sinus-Surgery datasets, we changed whether or not to combine the proposed methods with a segmentation model, focusing on the relative performance changes. We conducted 4 groups of ablation experiments. Group 1 and 2 were based on the two models described in Section III. Groups 3 was based on the model in Section IV.D. Group 4 was based on the re-implemented LWANet, which was featured by attention fusion block and depth-wise separable convolution [25]. In each group, three ablation experiments were conducted by controlling whether to use the proposed two methods. The evaluation results are shown in Table II. Several segmentation results are visualized in Fig. 4(a-d).

*1) Effectiveness of MAFA:* In each group, the comparison between the 1$^{st}$ and 2$^{nd}$ results showed the influence of MAFA. Over the four groups, the average mIOU improvements were 4.8% for Sinus-Surgery-C, and 7.6% for Sinus-Surgery-L. Therefore, MAFA provided a significant relative increase in segmentation accuracy.

According to the rotation-invariance metrics, we found that MAFA improved the rotational consistency of the segmentation models. Without MAFA, the average

TABLE II
ABLATION EXPERIMENT RESULTS WITH SINUS-SURGERY DATASETS AND DIFFERENT MODELS

| No. | Model | Proposed MAFA | Proposed CS | Sinus-Surgery-C mIOU (%) | Sinus-Surgery-C mRM$_{IOU}$ (%) | Sinus-Surgery-C mRSD$_{IOU}$ (%) | Sinus-Surgery-C mIOU$_{NB}$ (%) | Sinus-Surgery-L mIOU (%) | Sinus-Surgery-L mRM$_{IOU}$ (%) | Sinus-Surgery-L mRSD$_{IOU}$ (%) | Sinus-Surgery-L mIOU$_{NB}$ (%) | Time (ms) |
|---|---|---|---|---|---|---|---|---|---|---|---|---|
| 1.1 | DeepLabv3+ [28] with ResNet50 | × | × | 84.1±2.7 | 73.5±2.0 | 22.9±2.2 | 77.3±4.3 | 76.4±5.0 | 65.4±8.1 | 23.1±4.3 | 69.7±7.2 | 7.4 |
| 1.2 |  | √ | × | **87.7**±1.6 | ***88.0***±1.3 | ***2.4***±0.2 | **81.2**±2.4 | 81.9±5.3 | 80.9±4.9 | 5.3±1.1 | 74.2±6.8 | 14.1 |
| 1.3 |  | √ | √ | 87.1±1.4 | 87.5±1.1 | 2.5±0.3 | 81.0±2.6 | **83.9**±4.3 | **82.9**±4.2 | **4.4**±0.9 | **76.3**±5.6 | 14.1 |
| 2.1 | TernausNet [21] with VGG16 | × | × | 82.2±3.6 | 72.1±1.9 | 25.0±1.7 | 76.3±5.4 | 75.4±4.7 | 68.2±7.1 | 20.1±3.5 | 69.3±7.3 | 10.6 |
| 2.2 |  | √ | × | **87.3**±1.0 | 87.0±0.5 | 4.0±0.2 | ***81.6***±2.8 | 81.1±4.3 | 77.6±5.2 | 10.9±2.6 | 74.7±6.5 | 30.6 |
| 2.3 |  | √ | √ | 87.2±0.9 | **87.5**±0.7 | **3.4**±0.5 | 81.3±2.4 | **84.2**±3.4 | **83.0**±3.8 | **6.1**±1.5 | **78.3**±5.2 | 30.6 |
| 3.1 | DeepLabv3+ [28] with MobileNet | × | × | 80.0±3.1 | 70.6±1.0 | 22.9±1.4 | 72.0±5.2 | 69.9±4.9 | 63.3±6.6 | 22.2±3.4 | 62.1±6.1 | 3.2 |
| 3.2 |  | √ | × | 85.0±0.9 | 84.9±0.4 | 3.6±0.6 | 77.8±2.2 | 77.9±4.8 | 77.9±4.6 | 5.9±0.9 | 68.6±6.1 | 6.2 |
| 3.3 |  | √ | √ | **85.5**±0.8 | **85.8**±0.7 | **3.3**±0.6 | **78.9**±2.5 | **82.1**±3.6 | **81.2**±3.3 | **5.1**±0.8 | **73.4**±4.8 | 6.2 |
| 4.1 | LWANet [25] with MobileNet | × | × | 77.3±4.7 | 66.8±2.5 | 25.9±2.0 | 69.6±6.6 | 65.9±9.0 | 59.7±9.7 | 22.4±3.5 | 60.7±8.1 | 4.3 |
| 4.2 |  | √ | × | 82.9±1.7 | 82.0±1.3 | 5.7±0.6 | 74.9±3.2 | 77.2±5.8 | 75.8±5.6 | 8.3±1.9 | 70.9±5.2 | 8.0 |
| 4.3 |  | √ | √ | **85.1**±1.0 | **84.3**±0.7 | **4.7**±1.0 | **78.7**±3.1 | **81.0**±3.6 | **79.1**±4.0 | **3.6**±1.5 | **74.0**±4.8 | 8.0 |

* Bold and italic fonts indicate the best performance in the group and in the column, respectively.

|mIOU-mRM$_{IOU}$| was as high as 10.1% for Sinus-Surgery-C, and 7.7% for Sinus-Surgery-L, which meant that image rotation caused significant deviations between the results given by the un-rotated input and the rotated inputs, respectively. Using MAFA, the average |mIOU-mRM$_{IOU}$| was reduced to 0.4% for Sinus-Surgery-C, and 2.0% for Sinus-Surgery-L. Besides, MAFA reduced mRSD$_{IOU}$, indicating the better rotation-invariance performance. As shown in Fig. 4(a-d), rotations could lead to significant change of segmentation accuracy, when MAFA was not used.

*2) Effectiveness of contour supervision (CS)*: In each group, the comparisons between the 2$^{nd}$ and 3$^{rd}$ experiments showed the effectiveness of contour supervision. Over the four groups, the average mIOU improvements were 0.5% for Sinus-Surgery-C, and 3.3% for Sinus-Surgery-L. Therefore, the contour supervision provided a further relative increase of segmentation accuracy, except the two cases in Group 1 and 2 with Sinus-Surgery-C.

According to the metric mIOU$_{BN}$, it was shown that the segmentation accuracy near the instrument boundary was improved by contour supervision. Comparing the 2$^{nd}$ and 3$^{rd}$ results in each of the four groups, the average IOU$_{BN}$ improvement was 1.1% for Sinus-Surgery-C, and 3.4% for Sinus-Surgery-L. As shown in Fig. 4(a-d), the contour prediction presented good accuracy and sharpness, but might have incompleteness at the partial borders with low contrast.

*F. Comparison Experiments with EndoVis2017 Dataset*

The proposed methods were evaluated by the binary segmentation task on another public dataset provided by MICCAI 2017 Endoscopic Vision Challenge (EndoVis2017) [31]. Following [22], we used the same dataset with that in [22], containing 1350 training images and 450 testing images, whose resolution was 1280×1024.

DeepLabV3+ and the training configuration described in Section IV.B-E were adapted to this experiment with four modifications: 1) Considering the computation ability, the raw image was resized to 640×512 before inputted to the model. After inference, the segmentation map was resized back to 1280×1024. 2) During training, the batch size was 8 and dropout was not used. 3) The atrous rates in ASPP were set as {4,8}, because of the larger input size. The final concatenation-and-convolution operation in ASPP was replaced by the simpler summation operation. The feature channel numbers $N_{high}$ and $N_{low}$, which were introduced in Section III.A, were reduced down to 128 and 4, respectively. 4) We empirically deleted all the batch normalization layers except those in the backbone. The evaluation results are given in Table III. The three metrics were mDSC, mean *specificity* (mSpec.), and mean *sensitivity* (mSens.).

DeepLabV3+ with ResNet50 showed the highest specificity score, but its accuracy score mDSC was lower than that with MobileNet. Comparing rows 2 and 4, we found that MAFA could significantly improve the segmentation performance. MAFA ($N_A$=4) increased the mDSC, mSpec. and mSens. by 2.7%, 0.3% and 2.0%, respectively, compared to that without MAFA. Similar to mRSD$_{IOU}$, we used mRSD$_{DSC}$ here to indicate the rotation-invariance performance. For the models without MAFA, with MAFA ($N_A$=2), and with MAFA ($N_A$=4), the resulting mRSD$_{DSC}$ scores were 3.7%, 3.0%, and 2.3%, respectively, which showed that MAFA could reduce the rotational variance. In Fig. 4(e, f), two segmentation cases are given, which showed that rotation caused the larger inconsistency to the model without MAFA.

Comparing row 5 with rows 6-10, it is shown that DeepLabV3+ (MobileNet) with MAFA ($N_A$=4) surpassed the other existing segmentation models. Contour supervision did not provide significant performance improvement for binary segmentation on this dataset. This might be due to the stronger

TABLE III
COMPARISON EXPERIMENTS ON BINARY SEGMENTATION
WITH DEEPLABV3+ (DL3+) AND ENDOVIS2017 DATASET

| Method | | mDSC (%) | mSpec. (%) | mSens. (%) | Time (ms) |
|---|---|---|---|---|---|
| 1. DL3+ [28] (ResNet50) | | 89.8 | **99.2** | 87.6 | 16.8 |
| DL3+ [28] (MobileNet) | 2. **no MAFA** | 90.4 | 98.7 | 92.4 | 6.3 |
|  | 3. **MAFA** $N_A$=2 | 91.7 | 99.0 | 92.9 | 11.2 |
|  | 4. **MAFA** $N_A$=4 | **93.1** | 99.0 | **94.4** | 20.7 |
| 5. MFF [22] | | 91.6 | 98.9 | 92.8 | 5.8 |
| 6. LinkNet [19] | | 90.6 | 98.9 | 92.0 | 4.1 |
| 7. ICNet [24] | | 88.2 | 98.6 | 89.1 | 9.1 |
| 8. UNet [18] | | 87.8 | 98.5 | 83.0 | 4.5 |
| 9. PSPNet [23] | | 83.1 | 99.0 | 78.8 | 16.3 |

* Bold fonts indicate the best performance in the column.

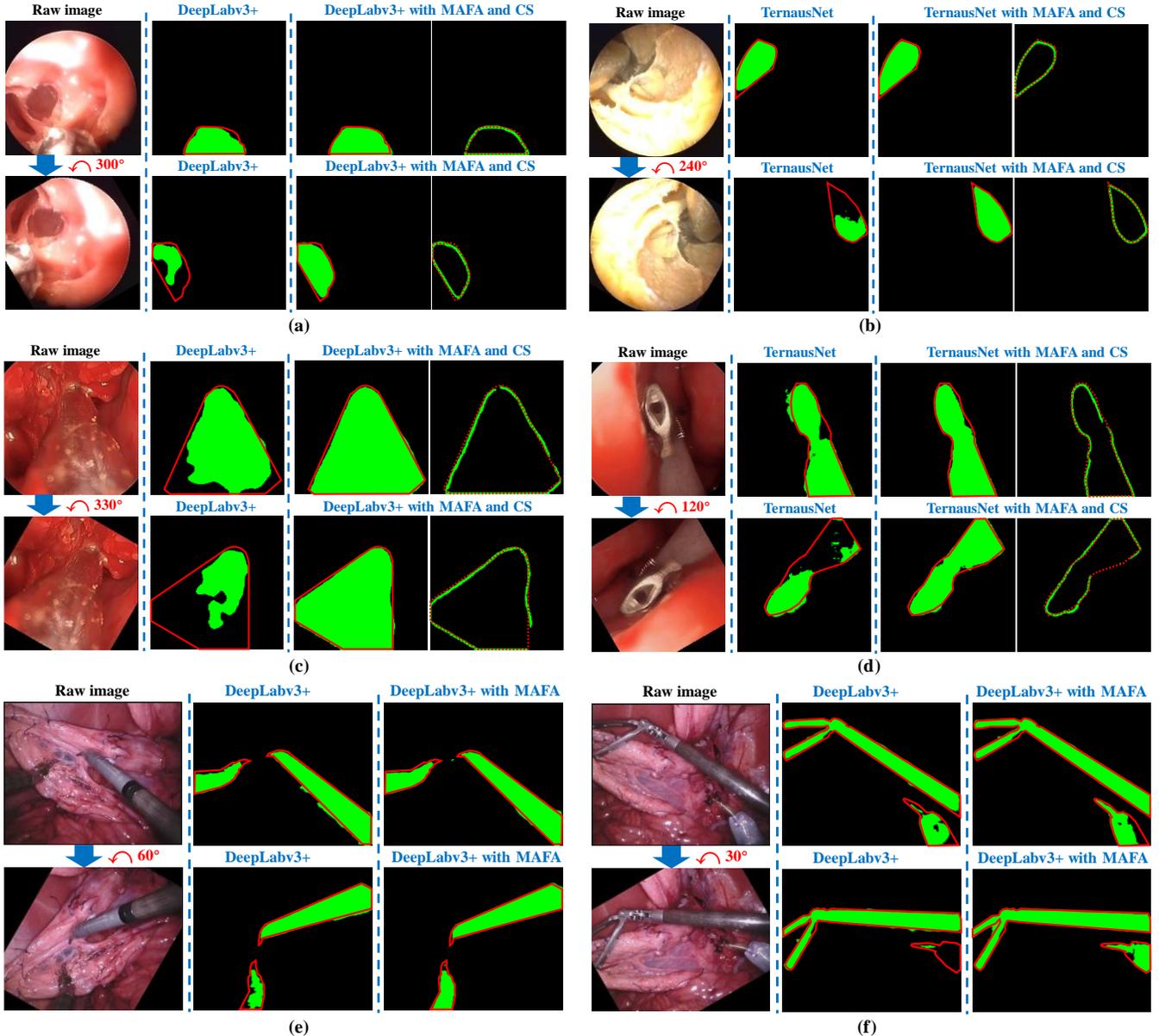

Fig. 4. Influence of image rotation on segmentation accuracy. In each sub-figure, the 1st and 2nd rows correspond to the raw image input and the rotated image input, respectively. The rotation angles are labeled by red digits. (a) and (b) are from Sinus-Surgery-C dataset. (c) and (d) are from Sinus-Surgery-L dataset. (e) and (f) are from EndoVis2017 dataset. The goundtruth of foreground is marked with red contour. The predicted foreground and contour are drawn in green. The backbone used in (a) and (c) is ResNet-50. The backbone used in (e) and (f) is MobileNet. $N_A$ of MAFA is 4.

instrument-background contrast in this dataset than in the Sinus-Surgery datasets, so that the contour awareness could be easily achieved without using auxiliary contour supervision.

### G. Discussion

MAFA provides augmented features by rotating the input image. Another way to augment features is to use a deeper and wider backbone feature extractor. First, a deeper and wider backbone will increase the parameter numbers for training but MAFA will not. Second, as shown in Table II, comparing the two lightweight models with MAFA to the two heavier models without MAFA, the former showed the better accuracy, and nearly the same or even faster speed. Therefore, the combination of MAFA and lighter model could surpass the heavier model with more layers and feature channels.

The contour prediction layer is a bypass of the segmentation layer. They are coupled by sharing the same input features. Therefore, although the contour supervision loss does not back-propagate through the segmentation layer, it guides the model to learn more contour aware features, which benefits the segmentation layer in an indirect manner.

## V. CONCLUSION

Aiming to improve surgical instrument segmentation in challenging endoscopic images, MAFA and contour supervision are proposed to enhance a deep segmentation model. The idea of MAFA is that by actively rotating an image with multiple angles, more visual cues can be collected. The semantic consistency is assumed under rotation-and-alignment, based on which the multi-angle features can be aligned in orientation and fused by summation. For better accuracy near the instrument boundary, the sharp contour prediction is added as an auxiliary learning task to guide the model to infer boundary-aware features. The proposed method can be flexibly combined with a deep model, without increasing the parameter number. The two novel instrument segmentation datasets were collected from endoscopic sinus surgeries, which have challenges caused by dexterous motion and narrow operating space. In the future, it is appealing to study the rotation-invariance behavior beyond the aspect of feature aggregation, and leverage temporal information across multi-frames to realize robust segmentation.